\documentclass[letterpaper]{article} 
\usepackage{aaai2026}  
\usepackage{times}  
\usepackage{helvet}  
\usepackage{courier}  
\usepackage[hyphens]{url}  
\usepackage{graphicx} 
\usepackage{natbib}  
\usepackage{caption} 
\usepackage{algorithm}
\usepackage{algorithmic}
\usepackage[hyphens]{url}  

\usepackage{newfloat}
\usepackage{listings}
\DeclareCaptionStyle{ruled}{labelfont=normalfont,labelsep=colon,strut=off} 
\floatstyle{ruled}
\newfloat{listing}{tb}{lst}{}
\floatname{listing}{Listing}

\usepackage{dblfloatfix}
\usepackage{graphicx}
\usepackage{subcaption}

\usepackage{amsmath}
\usepackage{booktabs} 
\usepackage{xcolor}
\newcommand{\answerYes}[1]{\textcolor{blue}{#1}} 
\newcommand{\answerNo}[1]{\textcolor{teal}{#1}} 
\newcommand{\answerNA}[1]{\textcolor{gray}{#1}}

\title{DisasterVQA: A Visual Question Answering Benchmark Dataset for Disaster Scenes}
\author {
    Aisha Al-Mohannadi,\textsuperscript{\rm 1,2}
    Ayisha Firoz,\textsuperscript{\rm 3}
    Yin Yang,\textsuperscript{\rm 2}
    Muhammad Imran,\textsuperscript{\rm 1}
    Ferda Ofli\textsuperscript{\rm 1}
}
\affiliations {
    \textsuperscript{\rm 1}~Qatar Computing Research Institute, Hamad Bin Khalifa University, Doha, Qatar\\
    \textsuperscript{\rm 2}~College of Science \& Engineering, Hamad Bin Khalifa University, Doha, Qatar\\
    \textsuperscript{\rm 3}~Department of Computer Science \& Engineering, Qatar University, Doha, Qatar\\
    aalmohannadi@hbku.edu.qa, af2400459@student.qu.edu.qa, yyang@hbku.edu.qa, mimran@hbku.edu.qa, fofli@hbku.edu.qa
}

\begin{document}

\maketitle

\begin{abstract}
Social media imagery provides a low-latency source of situational information during natural and human-induced disasters, enabling rapid damage assessment and response. While Visual Question Answering (VQA) has shown strong performance in general-purpose domains, its suitability for the complex and safety-critical reasoning required in disaster response remains unclear. We introduce DisasterVQA, a benchmark dataset designed for perception and reasoning in crisis contexts. DisasterVQA consists of 1,395 real-world images and 4,405 expert-curated question–answer pairs spanning diverse events such as floods, wildfires, and earthquakes. Grounded in humanitarian frameworks including FEMA ESF and OCHA MIRA, the dataset includes binary, multiple-choice, and open-ended questions covering situational awareness and operational decision-making tasks. We benchmark seven state-of-the-art vision–language models and find performance variability across question types, disaster categories, regions, and humanitarian tasks. Although models achieve high accuracy on binary questions, they struggle with fine-grained quantitative reasoning, object counting, and context-sensitive interpretation, particularly for underrepresented disaster scenarios. DisasterVQA provides a challenging and practical benchmark to guide the development of more robust and operationally meaningful vision–language models for disaster response. 
\end{abstract}

 \begin{links}
 \link{Dataset}{https://doi.org/10.5281/zenodo.18267769}
 \link{Code}{https://github.com/qcri/DisasterVQA}
 
 \end{links}

\section{Introduction}
\label{sec:intro}
Natural disasters such as floods, earthquakes, wildfires, and hurricanes cause widespread devastation, threatening lives, infrastructure, and economies. In their immediate aftermath, first responders and humanitarian agencies require timely and actionable intelligence to support response and recovery efforts. Rapidly understanding visual conditions on the ground is therefore critical for effective humanitarian operations. Social media plays an increasingly important role in modern disaster management by providing low-latency access to visual information from affected areas, provided that damage scenes can be accurately interpreted and systematically analyzed \cite{imran2020using, agarwal2020crisis}.

To this end, Vision-Language Models (VLMs), which integrate visual and textual information, offer promising capabilities for automating the analysis of complex scenes. Existing Visual Question Answering (VQA) benchmark datasets have enabled significant progress in understanding the everyday scenes \cite{goyal2017making, hudson2019gqa, al-tahan2024unibench, yu2024mm}. However, these datasets primarily focus on common objects, people, and daily activities, and do not capture the unique challenges posed by disaster imagery.

Some Earth observation-oriented VQA benchmarks address disaster-related scenarios using overhead satellite imagery, but they typically focus on narrow, task-specific objectives such as disaster prediction or building damage assessment, often framed within broader general-purpose evaluation settings \cite{danish2025geobench, soni2025earthdial, wang2025disasterm3}. In contrast, there is currently no VQA dataset that is \textit{(i)} specifically curated for disaster scenes depicted in natural, ground-level imagery and \textit{(ii)} explicitly grounded in established humanitarian disaster response frameworks. Disaster environments differ fundamentally from everyday settings: they are chaotic, partially damaged, and visually complex, requiring models to reason about critical conditions such as accessibility, damage severity, and safety risks. Without dedicated benchmarks, current VLMs cannot be systematically evaluated or effectively advanced for these high-stakes humanitarian applications.

\begin{figure*}[t]
  \centering
  \includegraphics[width=0.82\textwidth]{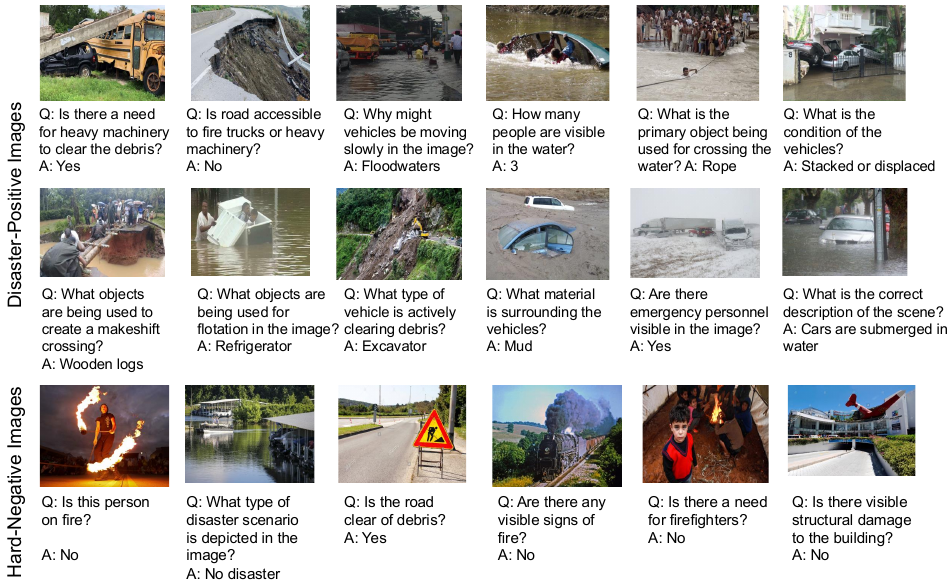}
  \caption{Sample image-question-answer triplets from the proposed DisasterVQA dataset}
  \label{fig:dataset-figure}
\end{figure*}

To tackle this challenge, we introduce DisasterVQA, the first large-scale VQA dataset centered on real-world natural disaster imagery. DisasterVQA consists of 4,405 carefully annotated image-question-answer triplets, covering a diverse range of disaster events. The questions are designed across three formats: binary, multiple-choice, and open-ended, to comprehensively assess VLMs' reasoning capabilities under varied levels of difficulty and ambiguity. Importantly, the questions are grounded in established humanitarian response frameworks (e.g., FEMA ESF~\cite{fema_esf}, OCHA MIRA~\cite{ocha_mira}, etc.) and cover tasks such as identifying built environment damage, determining hazard type and severity, evaluating accessibility, and assessing movement restrictions and controls,
making the dataset highly relevant for real-world disaster response applications. Figure~\ref{fig:dataset-figure} illustrates sample image-question-answer triplets from the dataset.

Alongside the dataset, we provide a comprehensive evaluation framework to drive reproducible research in disaster‑focused VQA, benchmarking seven state‑of‑the‑art VLM architectures (Molmo-7B, LLama-3.2-11B, Pixtral-12B, Mistral-Small-3.1-24B, Qwen2.5-VL-32B, GPT-4o-mini, GPT-4.1-mini) across all 
question types. 

The evaluation reveals that while binary questions are handled with relatively high reliability especially by GPT-4.1-mini, multiple-choice and open-ended questions remain challenging, with performance varying across disaster types and geographical regions. Open-weight models, though competitive in some areas, often show greater inconsistency. These results underscore the importance of region- and task-specific benchmarking in VQA, and highlight the need for improved visual reasoning in zero-shot settings.  By exposing these strengths and limitations, our benchmark offers actionable insights for advancing VLMs in high-stakes, real-world disaster scenarios.

\section{Related Work}
\label{sec:related_work}

VQA for disaster response remains underexplored. The only dedicated benchmark, FloodNet \cite{rahnemoonfar2021floodnet}, provides ~3,200 UAV images and 11,000 QA pairs for flood detection and road passability, but is restricted to aerial flood imagery with mostly binary and counting questions. Advanced prompting extensions \cite{sun2024reducing,karimi2025zeshot} are similarly constrained by its narrow scope. 

RSVQA \cite{lobry2020rsvqa} and its extension Prompt-RSVQA \cite{chappuis2022prompt} introduce VQA on satellite imagery for land cover and object counting, but rely on aerial views and standard environmental monitoring questions rather than the high-stakes reasoning required for disaster response. TAMMI \cite{boussaid2025visual} offers multimodal geospatial QA across RGB, multispectral, and SAR imagery, yet omits questions on accessibility, damage severity, or safety. Generic benchmarks, VQA v2.0 \cite{goyal2017making}, GQA \cite{hudson2019gqa}, and Visual7W \cite{zhu2016visual7w}, cover everyday scenes whose clean, object-centric images fail to capture the occlusions, debris, and visual chaos of disaster environments. 

Beyond VQA, xBD \cite{gupta2019creating} compiles ~850K building annotations from satellite imagery for segmentation and classification, but lacks the question-answer interface needed for interactive infrastructure or rescue reasoning. In summary, no existing benchmark provides multi-disaster, ground-level image QA pairs grounded in humanitarian frameworks. DisasterVQA addresses this gap directly.

\begin{figure*}[t]
  \centering
  \includegraphics[width=1\textwidth]{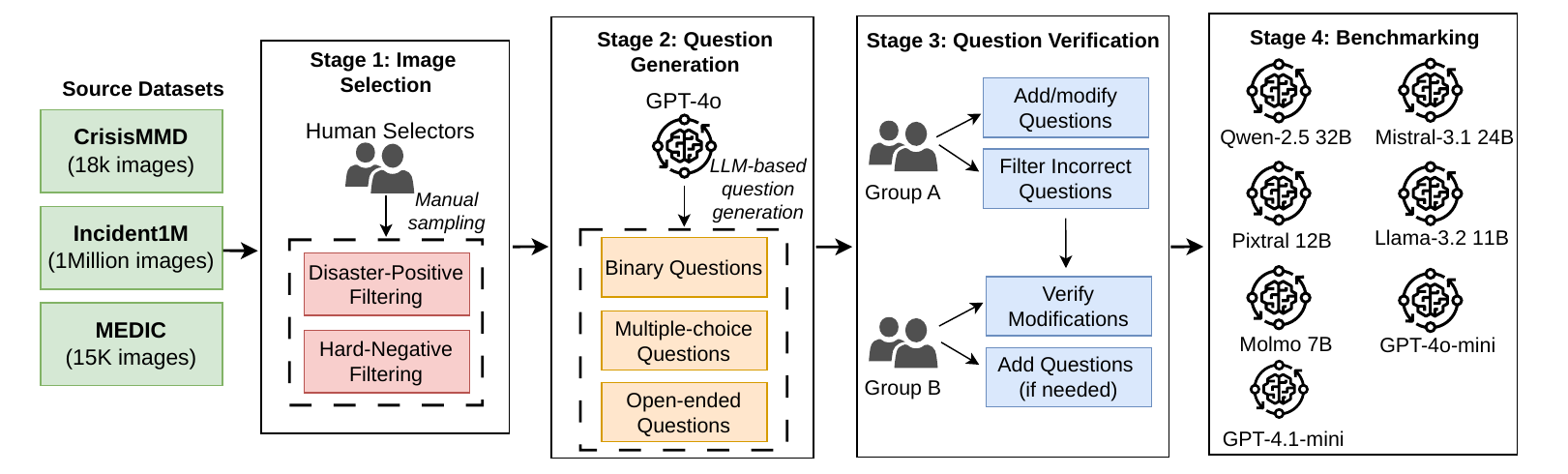}
  \caption{Overview of the four stages followed for the dataset curation and benchmarking}
  \label{fig:dataset-pipeline}
\end{figure*}

\section{DisasterVQA Construction}


The primary objective of DisasterVQA is to establish a benchmark that rigorously evaluates the capability of VLMs to function as reliable, operational assistants in crisis environments. Unlike general-domain VQA tasks, disaster response requires models to interpret chaotic visual evidence through specific humanitarian lenses to identify not just the presence of damage, but its nature, severity, and implications for response operations. Therefore, our construction process has two overarching goals: technical rigor and operational relevance. To achieve these goals, we employed a multi-stage, human-in-the-loop pipeline (as shown in Figure~\ref{fig:dataset-pipeline}) that integrates diverse source imagery with expert-verified annotations, as detailed in the following subsections.

\subsection{Image Selection}
We sourced our base imagery from three established datasets: CrisisMMD \cite{crisismmd2018icwsm}, Incidents1M \cite{weber2022incidents1mlargescaledatasetimages}, and MEDIC \cite{alam2022medic}.\footnote{Incidents1M is released under MIT License while CrisisMMD and MEDIC are available under CC BY-NC-SA 4.0 license.} While these repositories provide extensive coverage, they vary significantly in metadata quality and visual relevance. To ensure the dataset supports complex reasoning rather than simple classification, we implemented a targeted manual curation process executed by two human annotators. The selection strategy prioritized semantic density over volume, adhering to two strict criteria:

\begin{itemize}
\item \textbf{Disaster-positive images:} Annotators selected only those images exhibiting high visual complexity and multi-object scenes clearly indicative of disaster events (e.g., floods, wildfires, earthquakes). Images with low visual signal or minimal reasoning value were discarded to ensure every positive sample posed a genuine interpretative challenge.

\item \textbf{Hard-negative images:} To evaluate model resistance to false positives, we explicitly curated ``hard-negative" samples, which are non-disaster images that share visual textures or chaotic elements as disaster scenes (e.g., fireworks) The last row of Figure~\ref{fig:dataset-figure} shows more examples.
\end{itemize}

This yields 1,605 disaster-positive and 21 hard-negative images for the subsequent question-generation phase.

\begin{figure*}[t]
    \centering
    \begin{subfigure}[t]{0.28\textwidth}
        \centering
        \includegraphics[width=\linewidth]{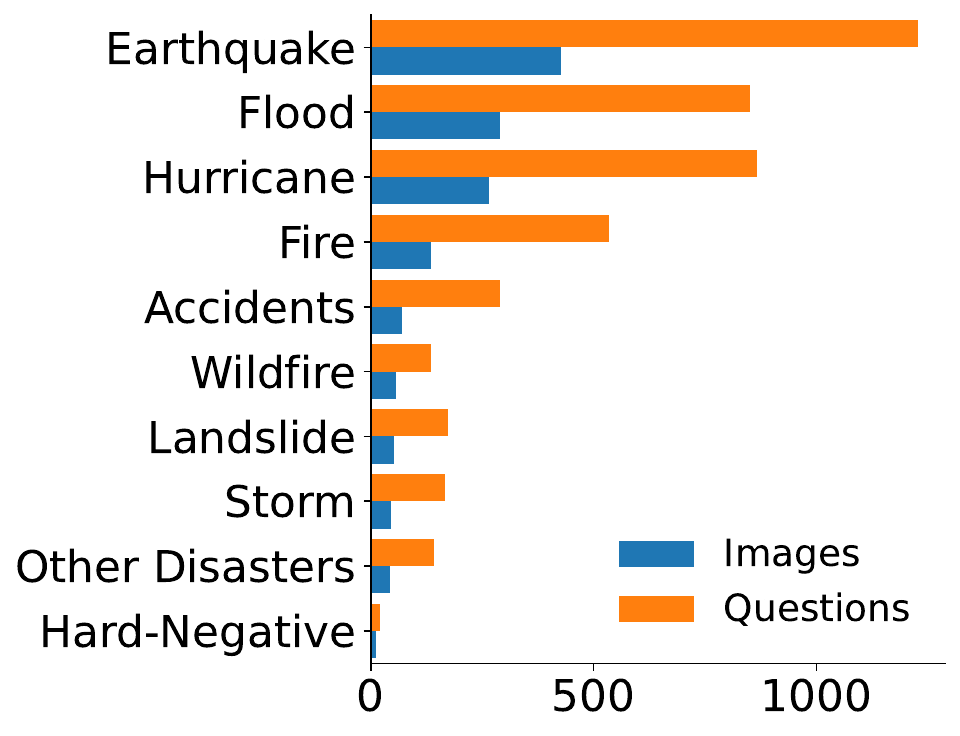}
        \caption{Distribution by disaster type}
        \label{fig:disaster-type}
    \end{subfigure}
    \hfill
    \begin{subfigure}[t]{0.28\textwidth}
        \centering
        \includegraphics[width=\linewidth]{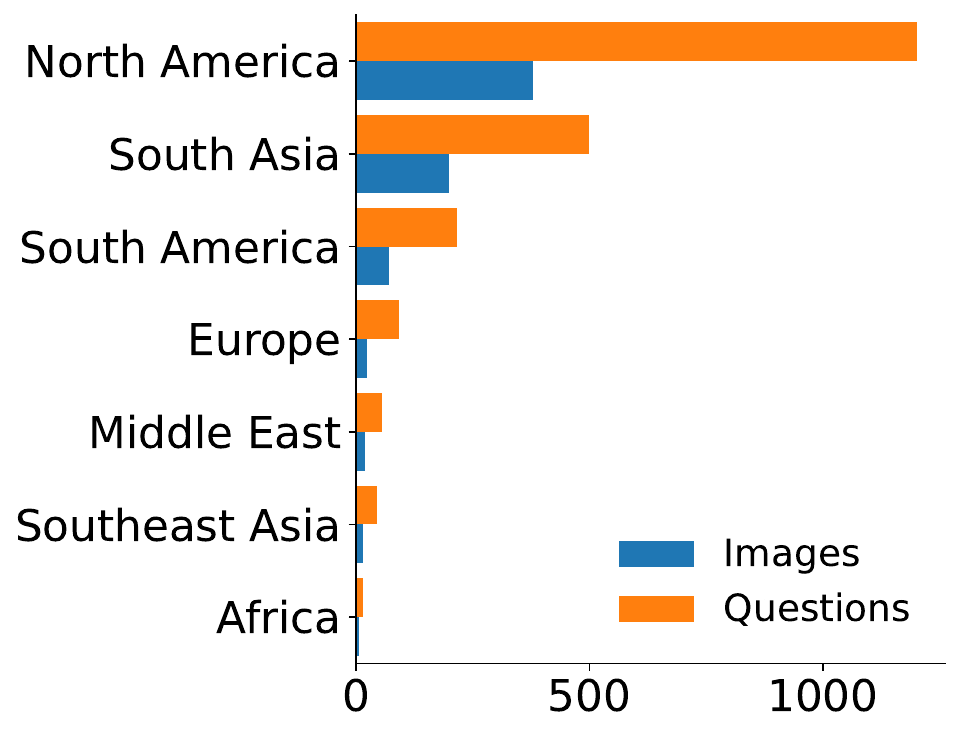}
        \caption{Distribution by geographic region}
        \label{fig:region}
    \end{subfigure}
    \hfill
    \begin{subfigure}[t]{0.40\textwidth}
        \centering
\includegraphics[width=\linewidth]{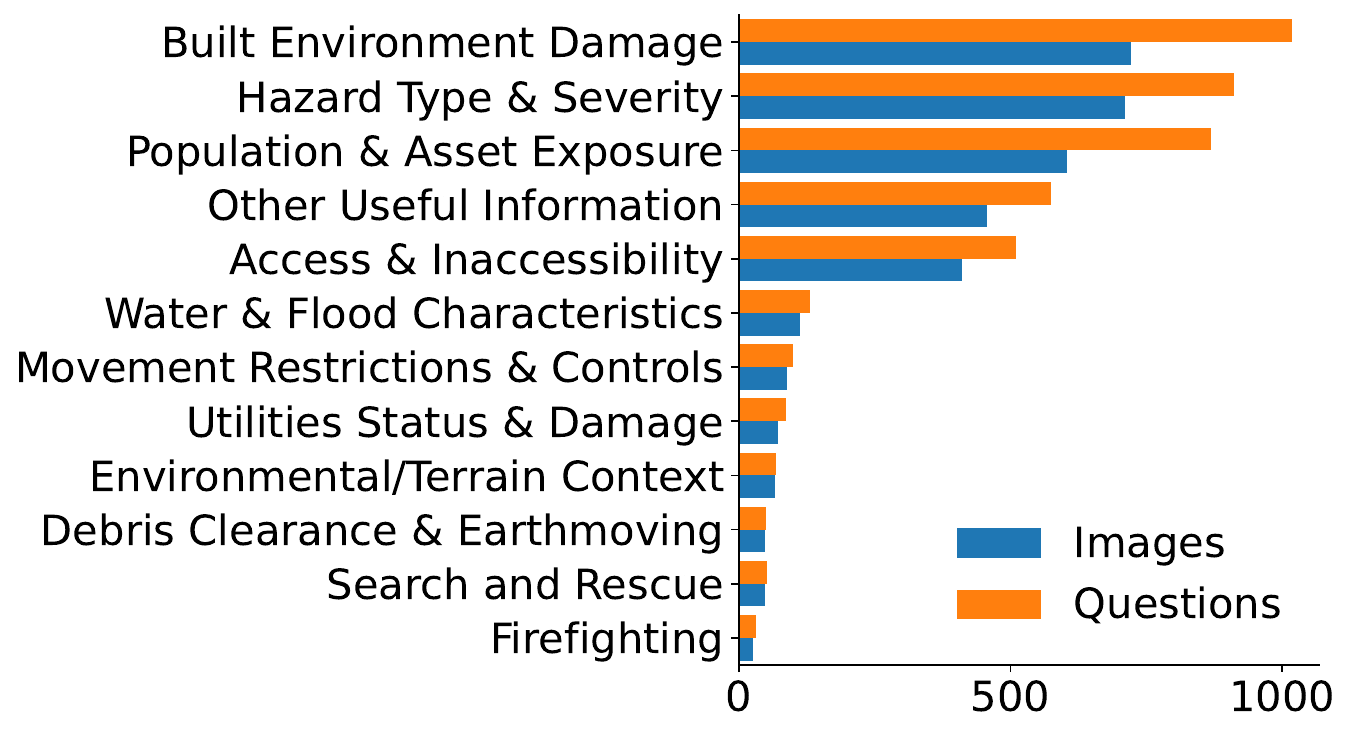}
        \caption{Distribution by humanitarian category}
        \label{fig:category}
    \end{subfigure}

    \caption{Overview of dataset composition in terms of image and question counts across (a) disaster types, (b) geographic regions, and (c) humanitarian categories.}
    \label{fig:dataset-overview}
\end{figure*}

\subsection{Question Generation} 
In this phase, we utilized GPT-4o to generate three distinct types of questions for each image: binary (yes/no), multiple-choice (MCQ), and open-ended. For each question type, the model was prompted to generate three unique questions, resulting in a total of nine questions per image. During this process, we encountered a few challenges, including issues related to image sensitivity and content appropriateness, which led to the exclusion of certain samples from the dataset. Consequently, this automated curation process refined the disaster-positive pool from 1,605 to 1,383 images, and the hard-negative set from 21 to 12 images. Next, we examined the candidate 12,555 Q\&As generated for 1,395 images (three Q\&As for each question type per image) to perform the human verification.

\subsection{Question Verification}
After question generation, we focused on question refinement and selection to guarantee question quality and appropriate difficulty, as we conducted two rounds of human question selection:
\begin{itemize}
    \item \textbf{Group A} is a team of two annotators who reviewed the raw GPT-4o output for triviality and correctness. They also corrected the model's incorrect answers and added new questions to address identified gaps. 
    \item \textbf{Group B}, consisting of two senior researchers with over a decade of experience in AI for disaster response, performed the final quality assurance pass. This group focused on semantic consistency and domain specificity, retaining questions that posed meaningful challenges for humanitarian analysis. Images yielding low-quality or irrelevant Q\&A pairs were discarded at this stage. 
\end{itemize}
The final human-verified dataset contains 4,405 question–answer pairs, including 2,153 binary (yes/no), 1,693 multiple-choice, and 559 open-ended questions.


\subsection{Humanitarian Mapping}
Next, we map our dataset to established humanitarian needs taxonomies. We employed GPT-4o to categorize each verified Q\&A pair based on definitions from several frameworks, including the FEMA Emergency Support Functions (ESF)~\cite{fema_esf} and OCHA MIRA~\cite{ocha_mira}. We instructed the model to classify questions into two operational themes: Situational Awareness and Actionable Information. Any valid disaster-related questions falling outside these specific schemas were preserved in a broader ``Other Useful Information" category.

\subsection{Final Dataset Statistics}
The final DisasterVQA dataset\footnote{The DisasterVQA dataset is available at \url{https://zenodo.org/records/18267770}.} comprises 1,395 images paired with 4,405 questions, representing a diverse array of crisis scenarios. As shown in Figure~\ref{fig:dataset-overview}.a, the data exhibit a naturally long-tailed distribution, reflecting real-world content availability pattern on social media. Earthquakes (427 images, 1,227 questions), Floods (289 images, 851 questions), and Hurricanes (265 images, 867 questions) dominate the corpus, together accounting for approximately 70\% of all samples. This skew toward large-scale natural hazards mirrors source distributions in CrisisMMD and Incidents1M, where globally visible events generate disproportionate media coverage. At the same time, we include localized and object-centric incident images, such as Fires (136 images, 534 questions) and Accidents (71 images, 289 questions), to evaluate model robustness beyond highly contextual disaster scenes. The Hard-Negative category contains 12 adversarial images paired with 21 questions and serves as a control set for assessing hallucination and overconfidence.

As most images are from specific disaster events, we leverage event information to infer country-level geolocation. The dataset spans a broad set of regions (as shown in Figure~\ref{fig:dataset-overview}.b), with strong representation from South Asia and North America, reflecting population density and reporting bias in social media–derived data. A substantial portion of samples (682 images with 2,279 questions), primarily sourced from Incident1M, are labeled as N/A due to missing or ambiguous location information. 
From an operational perspective, the dataset emphasizes information needs critical to humanitarian response (ref. Figure~\ref{fig:dataset-overview}.c): Built Environment Damage (722 images, 1,018 questions) and Hazard Type \& Severity (710 images, 912 questions) are the most prevalent categories, followed by Population/Asset Presence \& Exposure (604 images, 870 questions). Access \& Inaccessibility, central to routing and logistics, accounts for 410 images paired with 511 questions, while less frequent categories such as Utilities Status \& Damage, Search and Rescue, and Debris Clearance \& Earthmoving capture high-impact, mission-critical decision-making scenarios.



\section{Benchmarking Experiments}
\label{sec:benchmarking}

We evaluate a suite of VLMs on DisasterVQA to assess their operational perception and reasoning capabilities for disaster response. We benchmark seven state-of-the-art VLMs spanning diverse architectures, training paradigms, and model scales, including both strong open-weight and high-performing proprietary models. To ensure reproducibility and fair comparison, all experiments are conducted with the temperature parameter set to zero, yielding deterministic outputs. This systematic evaluation highlights the strengths and limitations of current VLMs in disaster contexts and establishes reference baselines to support future research.

\subsection{Vision-Language Models}
We evaluate the following seven state-of-the-art VLMs in our benchmarking experiments:

\begin{itemize}
    \item \textbf{Molmo-7B-D}~\cite{Deitke_2025_CVPR} is a 7B-parameter open-weight VLM that integrates a Qwen2-7B decoder with an OpenAI CLIP vision backbone via lightweight cross-modal adapters. The model is trained on PixMo, a corpus of one million carefully curated image-text pairs, and is optimized for multimodal pointing, captioning, and visual question answering tasks.
    \item \textbf{LLaMA-3.2-11B}~\cite{meta_ai_llama3.2_2024} is an 11B-parameter open-weight VLM from the LLaMA-3.2-Vision family of pretrained and instruction-tuned models. Built upon the LLaMA-3.1 text-only backbone, it is optimized for visual recognition, image reasoning, image captioning, and general visual question answering.
    \item \textbf{Pixtral-12B}~\cite{agrawal2024pixtral} is an open-weight VLM built on the Mistral Nemo 12B text decoder and a 400M-parameter custom vision encoder trained from scratch. The model supports native-resolution and variable-aspect-ratio image inputs, enabling interleaved text-image reasoning and multi-image processing within a 128K-token context window. It is designed for complex multimodal tasks such as document question answering, chart understanding, and multi-turn visual dialogue.
    \item \textbf{Mistral-Small-3.1-24B}~\cite{mistral_small3.1_2025} is a 24B-parameter open-weight VLM built on Mistral Small 3, augmented with enhanced vision capabilities for robust image understanding within a 128K-token context window. Its instruction-tuned variant is optimized for agentic multimodal tasks such as document question answering, chart reasoning, and visual instruction following, achieving strong performance on multimodal benchmarks while remaining efficient enough for single-GPU deployment.
    \item \textbf{Qwen2.5-VL-32B}~\cite{Qwen2.5-VL} is a 32B-parameter open-weight VLM from the Qwen-2.5-VL family, designed for advanced multimodal understanding and reasoning. It excels at interpreting text-rich and structured visual content, supports long-context multimodal inputs, and enables agentic capabilities such as visual grounding, tool use, and precise object localization with structured outputs. In this study, we use an Activation-aware Weight Quantization (AWQ) 4-bit variant, reducing memory usage by up to fourfold with less than a 3\% accuracy drop.
    \item \textbf{GPT-4o-mini}~\cite{openai_gpt4o-mini_2024} is a proprietary model from OpenAI that has multimodal capabilities across text, vision, and audio within a 128K-token context window. Despite its compact design, it achieves performance close to GPT-4o on challenging vision and reasoning benchmarks, making it well suited for large-scale and resource-constrained deployments.
    \item \textbf{GPT-4.1-mini}~\cite{openai_gpt4.1-mini_2025} is a proprietary, efficiency-oriented variant of the GPT-4.1 family that preserves the full 1M-token context window and the model's enhanced capabilities in coding and instruction following, while substantially reducing per-query costs and increasing throughput relative to GPT-4o. It also supports multimodal inputs, making it a scalable option for high-context, multimodal reasoning tasks.
\end{itemize}

\subsection{Output Post-processing}
With the refined questions in place, we performed zero-shot inference across all models. 
Because model outputs are often inconsistently formatted, we applied an LLM-based post-processing step using GPT-4o to standardize responses across question types. This step extracts clean, structured outputs: binary questions are normalized to exactly \texttt{Yes} or \texttt{No}, while multiple-choice responses are converted into a JSON array containing the selected option(s). This normalization enables reliable and automated evaluation.

For open-ended questions, we adopt an LLM-as-a-judge approach with GPT-4o to assess answer correctness in addition to output normalization. The judge model compares the extracted response against the ground-truth answer, accounting for semantic equivalence, paraphrasing, and formatting variations, and produces a structured decision (\texttt{Correct} or \texttt{Incorrect}) along with the normalized answer. Responses judged as correct are incorporated into the evaluation, while incorrect cases are flagged for further error analysis.

Furthermore, to validate the reliability of the GPT-4o judge in evaluating open-ended responses, a human-in-the-loop verification was conducted across a stratified sample of 196 model-image pairs (28 unique samples for each of the seven subject models).The expert audit revealed an overall judge accuracy of 99.5\%, demonstrating that the post-processing extraction logic and semantic alignment are highly robust. Specifically, the judge achieved a perfect 100\% accuracy rate in interpreting the outputs of six out of the seven models (GPT-4.1-Mini, GPT-4o-Mini, Llama-3.2-11B, Mistral-Small, Molmo-7B-D, and Pixtral). A singular extraction discrepancy was identified within the Qwen-2.5-VL subset, resulting in a minor variance and a model-specific accuracy of 96.4\%.
These results confirm that the semantic alignment and normalization pipeline effectively minimize false negatives and positives in the evaluation of complex, open-ended disaster scenarios.

\subsection{Evaluation Metrics}
We measure three metrics in our evaluations, one for each question type:
\begin{itemize}
    \item \textbf{Binary accuracy}: The fraction of binary questions where the model's \texttt{Yes}/\texttt{No} prediction exactly matches the ground truth over the total number of predictions: 
    \begin{equation}
\mathrm{Acc}_{\text{binary}}
= \frac{\mathrm{TP} + \mathrm{TN}}
       {\mathrm{TP} + \mathrm{TN} + \mathrm{FP} + \mathrm{FN}}.
\end{equation}
    \item \textbf{Open-ended accuracy}: 
    The percentage of open-ended questions where the model's output is rated as \texttt{Correct} by the LLM-as-a-judge:
    \begin{equation}
\mathrm{Acc}_{\text{open}}
= \frac{1}{N_{\text{open}}}
  \sum_{i=1}^{N_{\text{open}}}
  J\bigl(\hat y_i, y_i),
\end{equation}
 where \(N_{\text{open}}\) is the total number of open-ended questions, \(y_i\) is the ground truth answer, \(\hat y_i\) is the model's output on image-question pair \(i\), and \(J\bigl(\cdot,\cdot\bigr)\) is the LLM-as-a-judge function which returns 1 if the judge rates the answer as \texttt{Correct} and 0 otherwise.

 \item \textbf{Multiple-choice \(F_1\)-score}: We evaluate all MCQs jointly as a multi-label classification task. For each question, we compute the number of true positives (TP), false positives (FP), and false negatives (FN) across predicted answer options. The overall \(F_1\)-score is then given by
 \begin{equation}
F_{1,\mathrm{mcq}}
= \frac{2 \,\mathrm{TP}}{2\,\mathrm{TP} + \mathrm{FP} + \mathrm{FN}}
\end{equation}
which concisely balances the model's ability to avoid false positives against its ability to capture all true positives across the full set.

\end{itemize}

\subsection{Computational Resources} 
All five open-weight models (i.e., Molmo-7B-D, Llama-3.2-11B, Pixtral-12B, Mistral-Small-3.1-24B, and Qwen-2.5-VL-32B) were benchmarked on our on-premise cluster equipped with NVIDIA A100 GPUs. In contrast, the two GPT variants (i.e., GPT-4o-mini and GPT-4.1-mini) were accessed via the Microsoft's Azure OpenAI Service.


\section{Experimental Results}

\begin{figure}[t]
    \centering
    \includegraphics[width=.95\linewidth]{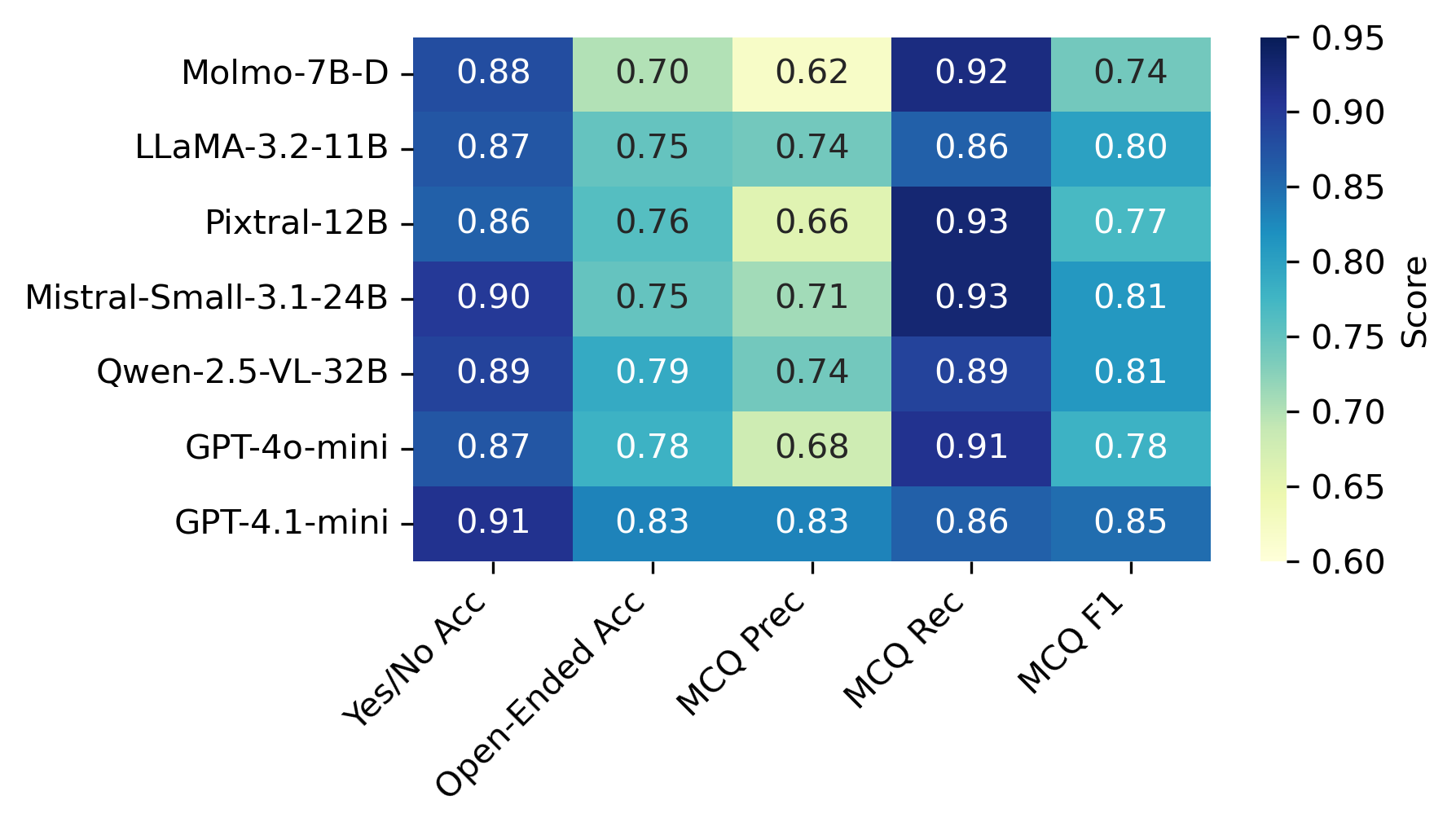}
    \caption{Overall performance of all the models for binary, open-ended, and multiple-choice questions}
    \label{fig:overall-heatmap}
\end{figure}

\begin{figure*}[t]
  \centering
  \includegraphics[width=.95\textwidth]{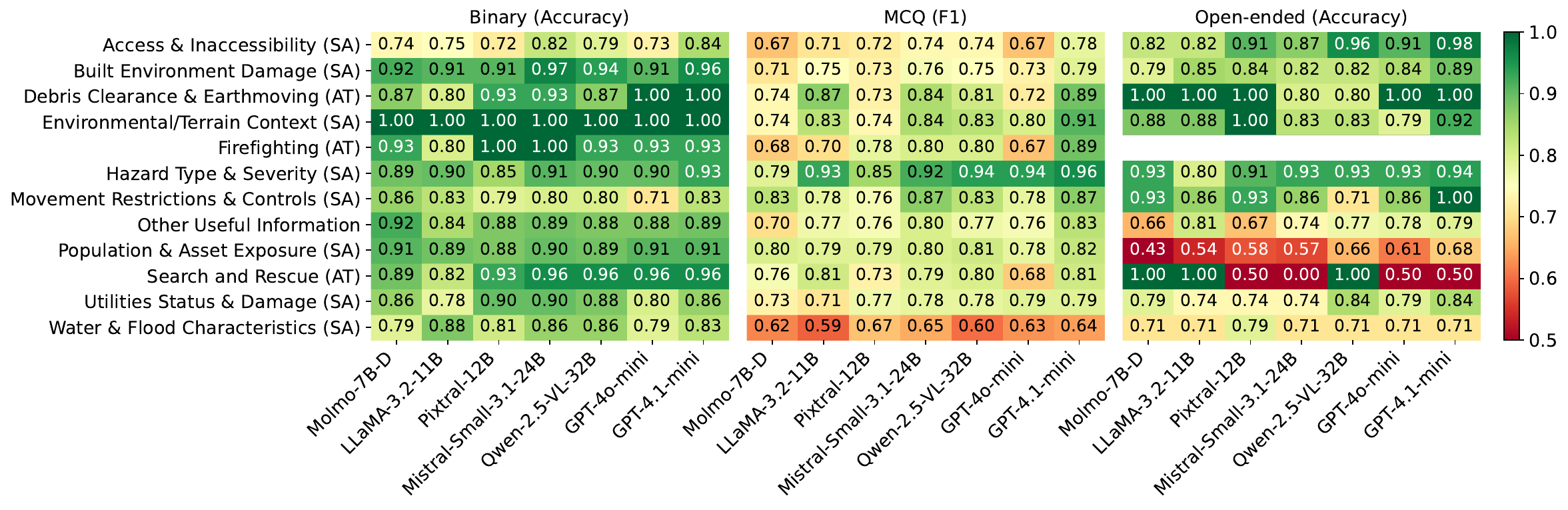}
  \caption{Performance by humanitarian categories across models. SA: Situational Awareness, AT: Actionable}
  \label{fig:hum_cats-heatmap}
\end{figure*}

In this section, we provide a detailed analysis of the experimental results across multiple evaluation dimensions. 
Figure~\ref{fig:overall-heatmap} summarizes model performance across the three question types. GPT-4.1-mini achieves the strongest overall results, with scores of 0.91 on binary questions, 0.83 on open-ended questions, and 0.85 on MCQs. Among open-weight models, Mistral-Small-3.1-24B and Qwen2.5-VL-32B emerge as close competitors, outperforming GPT-4o-mini on the majority of evaluations. Although Molmo-7B-D demonstrates competitive performance on binary questions, it lags behind other models on open-ended and multiple-choice questions, indicating limitations in more complex reasoning tasks.


\subsection{Analysis By Humanitarian Category}
Figure \ref{fig:hum_cats-heatmap} reports model performance across humanitarian categories. We evaluate each question format separately to show how performance changes from coarse detection (binary) to harder reasoning (open-ended and multiple-choice). Overall, binary results are strong, but gaps grow for categories that require fine-grained attributes (e.g., flood details) or counting/estimation (exposure).

\subsubsection{Binary Questions}
Binary accuracy is high across categories: \textit{Environmental/Terrain Context} is perfect (\(\approx 1.00\)) and \textit{Built Environment Damage} is consistently strong (\(\approx 0.93\)). The hardest binary category is \textit{Access \& Inaccessibility} (range \(\approx 0.72\)--\(0.84\)), with \textit{Movement Restrictions \& Controls} (\(\approx 0.80\)) and \textit{Water \& Flood Characteristics} (\(\approx 0.83\)) also below the top tier. Overall, the best binary model is \textbf{GPT-4.1-mini} (\(\approx 0.91\)), closely followed by \textbf{Mistral-Small-3.1-24B} (\(\approx 0.91\)); the top model varies by category (e.g., Mistral peaks on \textit{Built Environment Damage}, while LLaMA-3.2-11B peaks on \textit{Water \& Flood Characteristics}).

\subsubsection{Open-Ended Questions}
Open-ended accuracy varies more by category. \textit{Access \& Inaccessibility} is strongest (avg \(\approx 0.90\), best \(\approx 0.98\)), while \textit{Population \& Asset Exposure} is weakest (avg \(\approx 0.58\), down to \(\approx 0.43\)), consistent with counting and clutter challenges. \textit{Other Useful Information} is mid-range (avg \(\approx 0.71\)), and \textit{Search and Rescue} shows low average and high variance (avg \(\approx 0.64\), \(\approx 0.00\)--\(1.00\)), likely reflecting small coverage.

\subsubsection{Multiple-Choice Questions}
Multiple-choice performance (MCQ \(F_1\)) highlights fine-grained labeling difficulty. \textit{Hazard Type \& Severity} is strongest (avg \(F_1 \approx 0.90\), best \(\approx 0.96\)), while \textit{Built Environment Damage} is moderate (avg \(F_1 \approx 0.75\)), showing that detecting damage is easier than selecting the correct damaged assets. The weakest category is \textit{Water \& Flood Characteristics} (avg \(F_1 \approx 0.63\)), and \textit{Access \& Inaccessibility} is also challenging (avg \(F_1 \approx 0.72\)), where some models over-select options (higher false positives).

\subsubsection{SA vs.\ AT and Model Size}
Grouping categories shows that \textbf{AT} (Actionable) is easier than \textbf{SA} (Situational Awareness) for binary questions (SA \(\approx 0.88\) vs.\ AT \(\approx 0.92\)), mainly because difficult SA categories like \textit{Access \& Inaccessibility} and \textit{Movement Restrictions} require subtle “usability” judgments, while AT often has clearer visual evidence (responders, machinery, fire). Model size effects are modest for binary but clearer for structured reasoning: smaller open models average \(\approx 0.87\) on binary versus \(\approx 0.90\) for larger open models, with the biggest gaps in ambiguous SA categories; for open-ended and MCQ, larger models are generally more stable, especially on exposure (counting) and water (fine-grained attributes).

\begin{figure*}[t]
  \centering
  \includegraphics[width=.9\textwidth]{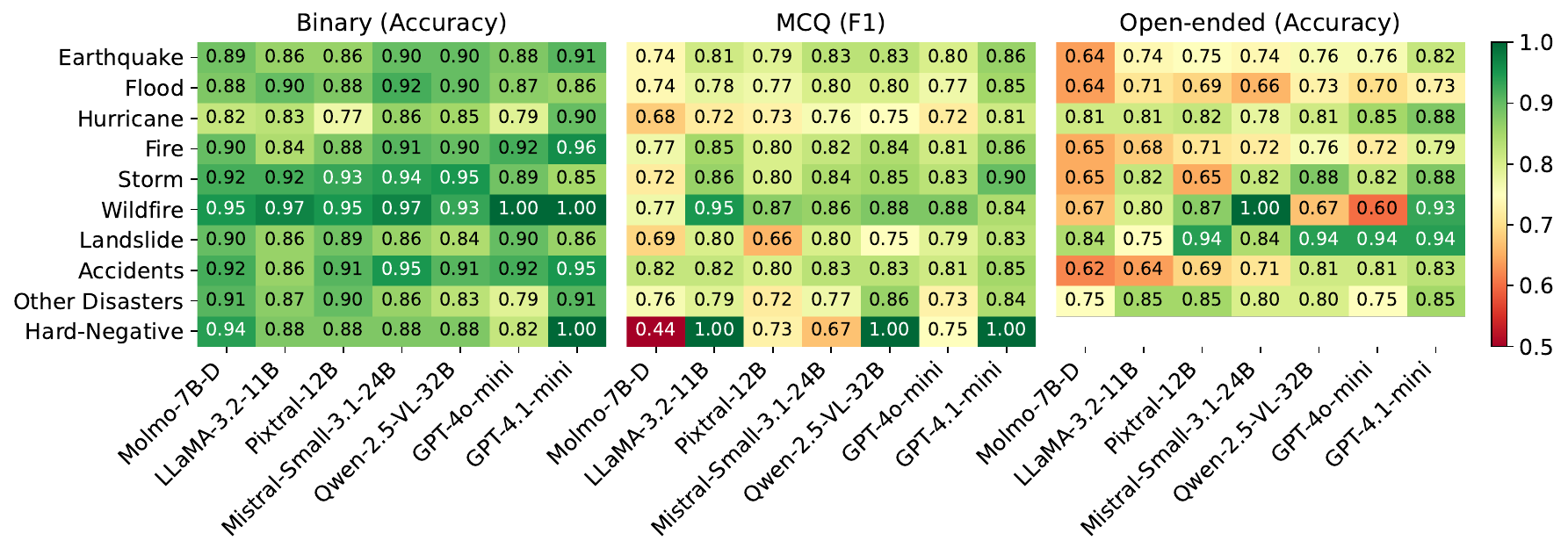}
  \caption{Models' performance across disaster types for binary, open-ended and multiple-choice questions}
  \label{fig:disaster-type-analysis}
\end{figure*}

\subsection{Analysis By Disaster Type}
Figure \ref{fig:disaster-type-analysis} illustrates the performance of the state-of-the-art models on different disaster types. We have evaluated each question type separately to observe the performance across different metrics. 

\subsubsection{Binary Questions} The accuracy is uniformly high: even the weakest case (Storm) remains at 0.85 with GPT-4.1-mini, and every other disaster is at least 0.86. Fire and Wildfire top out at 0.96–1.00, reflecting visually unambiguous damage cues, while Storms and Floods (0.85–0.86) prove slightly tougher. 

\subsubsection{Open-Ended Questions} the accuracy shows the largest spread, from a low of 0.73 on Floods to a peak of 0.94 on Landslides (GPT-4.1-mini). Accidents and Earthquakes sit in the low‐to‐mid 0.80s, whereas Fires drop to 0.79 and Hurricanes actually climb to 0.88.

\subsubsection{Multiple-Choice Questions} the \(F_1\) score falls between the two: Storms again lead with 0.90, followed by Earthquakes (0.86) and Floods/Accidents (0.85). The most challenging MCQs are for Hurricanes (0.81) and Landslides (0.83), followed by Wildfires (0.84), likely due to complex multi-object scenes and less‐straightforward choices.

\subsubsection{Hard-Negative}
Although the hard-negative subset is small, it is useful for checking false positives. Most top-performing models achieve perfect or near-perfect binary accuracy (e.g., GPT-4.1-mini at 1.00), while smaller models are slightly lower (e.g., Pixtral-12B at 0.88). In the multiple-choice format, performance is generally strong for the best models (near-perfect \(F_1\) for GPT-4.1-mini and Qwen2.5-VL), and Pixtral-12B remains competitive at \(F_1=0.73\). Overall, these results suggest that models can usually recognize non-disaster scenes, but the hard-negative set still exposes mild variability across smaller architectures.


\begin{figure*}[t]
  \centering
  \includegraphics[width=.8\textwidth]{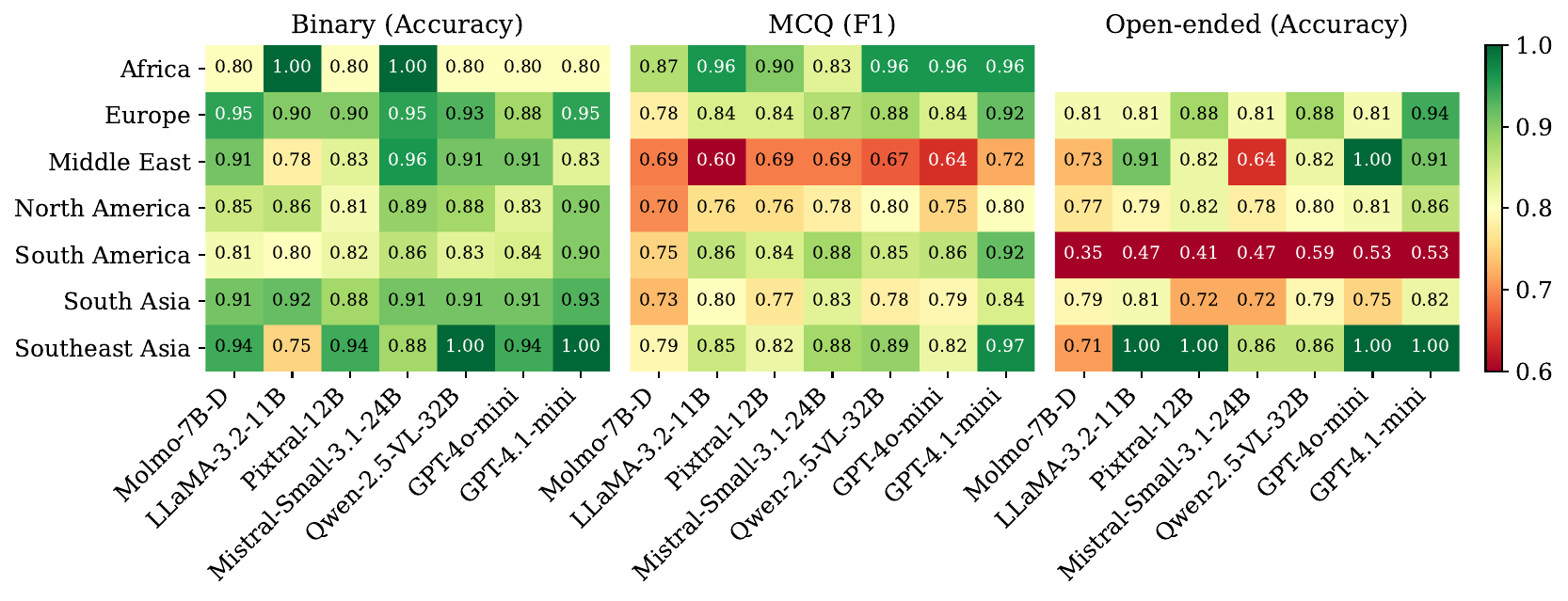}
  \caption{Region-based performance analysis for binary, open-ended, and multiple-choice questions}
  \label{fig:region-analysis-heatmap}
\end{figure*}

\subsection{Analysis By Region}
Figure \ref{fig:region-analysis-heatmap} presents three heatmaps of model performance across different geographic regions. There are several patterns that identify the strengths and weaknesses of these models, and we describe them in detail by analyzing each question type.

\subsubsection{Multiple-Choice Questions}
In Figure~\ref{fig:region-analysis-heatmap} (MCQ), the \textit{Middle East} is the hardest region: scores cluster around \(0.60\)–\(0.72\) (e.g., LLaMA-3.2-11B \(=0.60\), GPT-4.1-mini \(=0.72\)). In contrast, \textit{Africa} is the easiest, with all models \(\ge 0.83\) and several reaching \(0.96\). GPT-4.1-mini is the strongest overall, topping or tying every region, but its lowest value is in the Middle East (\(0.72\)). Open-source models are competitive in well-performing regions (e.g., South America and Southeast Asia), yet drop more sharply in the Middle East where all models compress toward the lower end of the scale.

\subsubsection{Binary Questions}
In Figure~\ref{fig:region-analysis-heatmap} (Binary), accuracy is high overall across regions, with the lowest value still at \(0.75\) (LLaMA-3.2-11B in Southeast Asia). However, regional differences remain visible: \textit{Africa} shows a split between perfect scores for some open models (LLaMA-3.2-11B and Mistral-Small at \(1.00\)) and lower scores for others (including GPT models at \(0.80\)). \textit{South Asia} is consistently strong across models (roughly \(0.88\)–\(0.93\)), while \textit{Middle East} is more mixed (e.g., LLaMA-3.2-11B \(=0.78\) vs.\ Mistral-Small \(=0.96\)). Notably, GPT-4.1-mini peaks at \(1.00\) in \textit{Southeast Asia} and remains solid elsewhere (typically \(0.83\)–\(0.95\)).

\subsubsection{Open-Ended Questions}
In Figure~\ref{fig:region-analysis-heatmap} (Open-ended), regional variation becomes much stronger than in binary questions. \textit{North America} is clearly the hardest region, with accuracies dropping to \(0.35\)–\(0.59\) across models, suggesting that free-form reasoning and extraction is sensitive to region-specific visual patterns and dataset composition. In contrast, \textit{South Asia} and \textit{Southeast Asia} are the easiest: several models reach perfect accuracy (\(1.00\)) in these regions, including GPT-4.1-mini and GPT-4o-mini. Europe and the Middle East remain strong for the top models (e.g., GPT-4.1-mini \(=0.91\) in Europe and \(0.86\) in the Middle East), while smaller open models show larger drops, especially in the lowest-performing regions.

\section{Discussion}

\subsection{Qualitative Error Analysis}
\begin{figure*}[t]
  \centering
    \includegraphics[width=.85\textwidth]{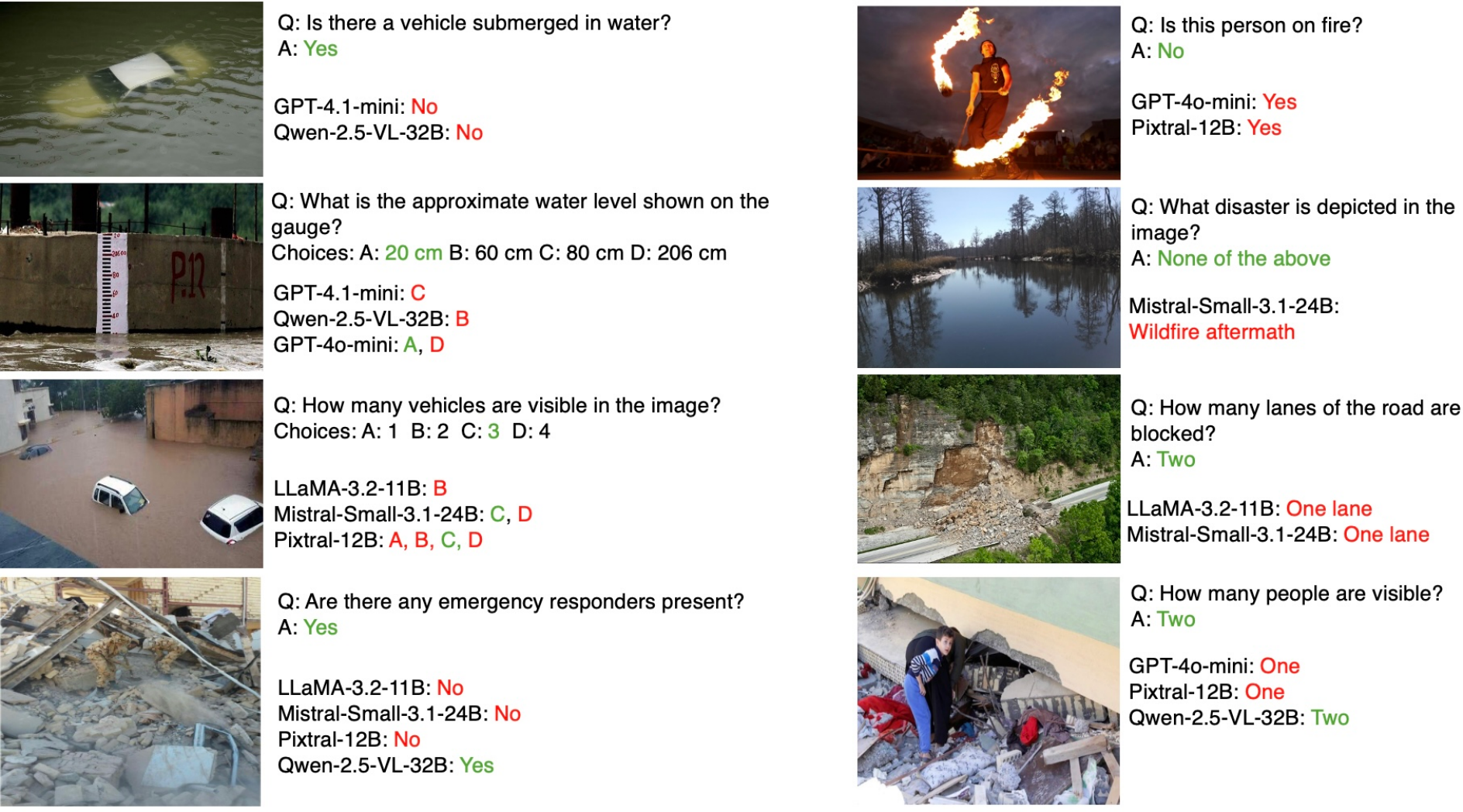}
    \caption{Models mistakes across various types of questions}
    \label{fig:errors}
\end{figure*}

Figure~\ref{fig:errors} illustrates representative failure cases across models, revealing several recurring error patterns. A common issue is missed visual evidence in partially occluded or low-salience regions. For instance, when asked whether a vehicle is submerged in water, both GPT-4.1-mini and Qwen-2.5-VL-32B incorrectly answered No despite a clearly visible submerged vehicle. This suggests sensitivity to contrast, viewpoint, or object completeness, where partially visible objects are discounted during reasoning.

A second dominant failure mode concerns fine-grained quantitative estimation and counting. Models frequently disagreed on multiple-choice questions, such as estimating water level from a gauge or counting the number of visible vehicles. In some cases, models produced multiple mutually exclusive answers (e.g., selecting several choices simultaneously), indicating uncertainty handling issues in structured question formats. Similar errors appear in infrastructure assessment tasks, where models underestimated the number of blocked lanes in landslide scenes, likely due to difficulty interpreting spatial extent and depth from a single viewpoint.

Finally, we observe semantic over-interpretation and hallucination, particularly in safety-critical contexts. In one example, GPT-4o-mini and Pixtral-12B incorrectly labeled a person performing fire dancing as “on fire,” conflating controlled flames with an emergency condition. Likewise, Mistral-Small-3.1-24B misclassified a calm flooded forest scene as “wildfire aftermath,” reflecting a bias toward disaster labeling even when visual cues do not support such conclusions. Collectively, these errors highlight that while models perform well on high-level recognition, they remain brittle in tasks requiring precise visual grounding, numerical reasoning, and restraint from overgeneralizing disaster semantics, capabilities that are critical for reliable humanitarian decision support.

\subsection{Ethical Considerations and Limitations}

DisasterVQA is designed as an evaluation benchmark for assessing vision–language perception and reasoning in disaster scenarios, rather than as a dataset for model training or deployment. Accordingly, its limitations primarily concern the scope and representativeness of the evaluation, not downstream operational use.

Furthermore, we acknowledge limitations regarding the dataset scale. 
DisasterVQA comprises 1,395 images and ~4,405 QA pairs, which is smaller than generic VQA benchmarks. However, because disaster imagery is inherently scarce and complex, we prioritized expert-curated, high-quality annotations strictly grounded in humanitarian frameworks over massive, noisy data scraping. Regarding methodological innovation, our dataset curation and evaluation pipelines rely heavily on existing LLMs, which is becoming standard practice. 
By adapting these automated pipelines to the complex disaster response domain and enforcing multi-stage human verification, we provide a robust testbed that reveals critical reasoning failures in current SOTA VLMs.

Because the dataset is constructed by sampling images from existing publicly available sources, it inevitably inherits their underlying biases, despite our careful curation and expert verification procedures aimed at mitigating these effects. In particular, social media–sourced imagery exhibits geographic, demographic, temporal, and device-related biases, with disproportionate representation of urban, well-connected regions and visually salient events. Platform-level amplification further skews coverage toward dramatic scenes, while subtler yet operationally important impacts may be underrepresented. These characteristics constrain how comprehensively DisasterVQA can probe model behavior across the full spectrum of real-world disaster conditions.
Moreover, DisasterVQA does not evaluate content authenticity or misinformation, which represents a distinct challenge requiring dedicated verification frameworks beyond the scope of this work.

Although DisasterVQA itself is not intended for operational use, the interpretation of benchmark results carries broader implications. Performance metrics should not be treated as indicators of deployment readiness, as errors in disaster-related perception and reasoning can have significant societal consequences. DisasterVQA is therefore best understood as a controlled diagnostic tool for identifying strengths, failure modes, and research gaps in current vision–language models, rather than as a proxy for real-world disaster response performance. 
To support responsible and transparent research use, DisasterVQA is released in alignment with the FAIR principles~\cite{fair} and accompanied by a Datasheet~\cite{gebru2021datasheets}. 

Researchers using DisasterVQA are expected to adhere to ethical data handling practices, avoid attempts at re-identification, and refrain from deploying models evaluated on this benchmark without appropriate privacy safeguards, human oversight, and complementary validation. The benchmark is provided to advance responsible AI research and should be interpreted strictly as an evaluation resource rather than a measure of real-world readiness.
\section{Conclusion}
\label{sec:conclusion}

In this work, we have introduced DisasterVQA, a rigorously curated visual‐question‐answering benchmark spanning nine disaster categories and eight state-of-the-art vision–language models. Through a multi-stage pipeline of image selection, automated question generation, human refinement, and LLM-based post-processing, we assembled 4,405 high-quality question–answer pairs over 1,395 images.
Our dataset captures a wide range of real-world crisis scenes and question types, including binary, multiple-choice, and open-ended, designed to reflect operational tasks such as damage assessment, accessibility evaluation, and hazard identification. We benchmarked seven prominent VLM 
models, and analyzed their performance across disaster types, question formats, humanitarian categories, and global regions. Results highlight strong zero-shot reasoning in binary questions, emerging capabilities in open-ended responses, and persistent challenges in geographically or visually ambiguous cases.
DisasterVQA serves as both a benchmarking resource and an analytical framework, uncovering limitations in current VLM performance and informing the design of more reliable, context-aware systems for real-world disaster response.

\bibliography{aaai2026}

\subsection{Paper Checklist}

\begin{enumerate}
\item For most authors...
\begin{enumerate}
    \item  Would answering this research question advance science without violating social contracts, such as violating privacy norms, perpetuating unfair profiling, exacerbating the socio-economic divide, or implying disrespect to societies or cultures?
    \answerYes{Yes}
  \item Do your main claims in the abstract and introduction accurately reflect the paper's contributions and scope?
    \answerYes{Yes}
   \item Do you clarify how the proposed methodological approach is appropriate for the claims made? 
    \answerYes{Yes}
   \item Do you clarify what are possible artifacts in the data used, given population-specific distributions?
    \answerYes{Yes}
  \item Did you describe the limitations of your work?
    \answerYes{Yes}
  \item Did you discuss any potential negative societal impacts of your work?
    \answerYes{Yes}
  \item Did you discuss any potential misuse of your work?
    \answerYes{Yes}
  \item Did you describe steps taken to prevent or mitigate potential negative outcomes of the research, such as data and model documentation, data anonymization, responsible release, access control, and the reproducibility of findings?
    \answerYes{Yes}
  \item Have you read the ethics review guidelines and ensured that your paper conforms to them?
    \answerYes{Yes}
\end{enumerate}

\item Additionally, if your study involves hypotheses testing...
\begin{enumerate}
  \item Did you clearly state the assumptions underlying all theoretical results?
    \answerNA{N/A}
  \item Have you provided justifications for all theoretical results?
    \answerNA{N/A}
  \item Did you discuss competing hypotheses or theories that might challenge or complement your theoretical results?
    \answerNA{N/A}
  \item Have you considered alternative mechanisms or explanations that might account for the same outcomes observed in your study?
    \answerNA{N/A}
  \item Did you address potential biases or limitations in your theoretical framework?
    \answerNA{N/A}
  \item Have you related your theoretical results to the existing literature in social science?
    \answerNA{N/A}
  \item Did you discuss the implications of your theoretical results for policy, practice, or further research in the social science domain?
    \answerNA{N/A}
\end{enumerate}

\item Additionally, if you are including theoretical proofs...
\begin{enumerate}
  \item Did you state the full set of assumptions of all theoretical results?
    \answerNA{N/A}
	\item Did you include complete proofs of all theoretical results?
    \answerNA{N/A}
\end{enumerate}

\item Additionally, if you ran machine learning experiments...
\begin{enumerate}
  \item Did you include the code, data, and instructions needed to reproduce the main experimental results (either in the supplemental material or as a URL)?
    \answerYes{Yes}
  \item Did you specify all the training details (e.g., data splits, hyperparameters, how they were chosen)?
    \answerYes{Yes}
  \item Did you report error bars (e.g., with respect to the random seed after running experiments multiple times)?
    \answerNo{No. We did not report error bars because all models were evaluated under deterministic settings (temperature set to zero), resulting in identical outputs across repeated runs. This design choice ensures reproducibility and enables fair, seed-independent comparisons between models.}
	\item Did you include the total amount of compute and the type of resources used (e.g., type of GPUs, internal cluster, or cloud provider)?
    \answerYes{Yes}
     \item Do you justify how the proposed evaluation is sufficient and appropriate to the claims made? 
    \answerYes{Yes}
     \item Do you discuss what is ``the cost`` of misclassification and fault (in)tolerance?
    \answerYes{Yes}
  
\end{enumerate}

\item Additionally, if you are using existing assets (e.g., code, data, models) or curating/releasing new assets, \textbf{without compromising anonymity}...
\begin{enumerate}
  \item If your work uses existing assets, did you cite the creators?
    \answerYes{Yes}
  \item Did you mention the license of the assets?
    \answerYes{Yes}
  \item Did you include any new assets in the supplemental material or as a URL?
    \answerYes{Yes}
  \item Did you discuss whether and how consent was obtained from people whose data you're using/curating?
    \answerNA{N/A}
  \item Did you discuss whether the data you are using/curating contains personally identifiable information or offensive content?
    \answerYes{Yes}
\item If you are curating or releasing new datasets, did you discuss how you intend to make your datasets FAIR (see \citet{fair})?
    \answerYes{Yes. We designed DisasterVQA with the FAIR principles in mind. The dataset will be Findable through a public repository with persistent identifiers and detailed metadata; Accessible via an open license and standard download mechanisms; Interoperable by using widely adopted data formats (e.g., JSON for image-question-answer triplets) and clearly defined schemas; and Reusable through comprehensive documentation, data curation guidelines, and evaluation protocols that enable reproducibility and downstream use.}
\item If you are curating or releasing new datasets, did you create a Datasheet for the Dataset (see \citet{gebru2021datasheets})? 
    \answerYes{Yes. A formal Datasheet for Datasets is provided in the official repository to document the collection, preprocessing, and ethical considerations of the DisasterVQA benchmark.}
\end{enumerate}

\item Additionally, if you used crowdsourcing or conducted research with human subjects, \textbf{without compromising anonymity}...
\begin{enumerate}
  \item Did you include the full text of instructions given to participants and screenshots?
    \answerNA{N/A}
  \item Did you describe any potential participant risks, with mentions of Institutional Review Board (IRB) approvals?
    \answerNA{N/A}
  \item Did you include the estimated hourly wage paid to participants and the total amount spent on participant compensation?
    \answerNA{N/A}
   \item Did you discuss how data is stored, shared, and deidentified?
   \answerNA{N/A}
\end{enumerate}

\end{enumerate}

\end{document}